\documentclass[journal]{IEEEtran}
\usepackage{booktabs}
\usepackage[table,xcdraw]{xcolor}

\usepackage{subfigure}
\usepackage[linesnumbered,ruled]{algorithm2e}
\usepackage{amsthm}
\usepackage{amsfonts}

\usepackage[cmex10]{amsmath}
\usepackage{hyperref}
\usepackage{threeparttable}
\usepackage[normalem]{ulem}

\usepackage{tabularx}
\usepackage{tabu}

\usepackage{tikz}
\usetikzlibrary{tikzmark}

\usepackage{graphicx}

\usepackage{amssymb}
\usepackage[]{algorithm2e}
\usepackage{graphicx}
\usepackage{booktabs}
\usepackage[table,xcdraw]{xcolor}
\usepackage{subfigure}
\usepackage{amsmath}
\usepackage{adjustbox}
\usepackage{verbatim}

\usepackage{float}

\usepackage{xcolor}
\usepackage[utf8]{inputenc}
\usepackage[english]{babel}
\usepackage{booktabs,caption,fixltx2e}

\usepackage{url}
\usepackage{ragged2e}

\ifCLASSOPTIONcompsoc
  \usepackage[nocompress]{cite}
\else
  \usepackage{cite}
\fi
\ifCLASSINFOpdf
\else
\fi
\makeatletter
\g@addto@macro{\endtabular}{\rowfont{}}
\makeatother
\newcommand{\rowfonttype}{}
\newcommand{\rowfont}[1]{
\gdef\rowfonttype{#1}#1\ignorespaces%
}
\makeatother

\begin{document}
%
\title{Multi-label Retinal Disease Classification Using Transformers}
%
%
%
%

\author{M. A. Rodr\'{i}guez, H. AlMarzouqi, and P. Liatsis
\IEEEcompsocitemizethanks{\IEEEcompsocthanksitem M. A. Rodr\'{i}guez, H. AlMarzouqi and P. Liatsis are with the Department of Electrical Engineering and Computer Science, Khalifa University, Abu Dhabi, United Arab Emirates.\protect\\
E-mail: \{100058256, hasan.almarzouqi,  panos.liatsis\}@ku.ac.ae
}
\thanks{}}

%
%

\markboth{}%
{Shell \MakeLowercase{\textit{et al.}}: Bare Demo of IEEEtran.cls for Computer Society Journals}
\IEEEtitleabstractindextext{%
\justify
\begin{abstract}
Early detection of retinal diseases is one of the most important means of preventing partial or permanent blindness in patients. In this research, a novel multi-label classification system is proposed for the detection of multiple retinal diseases, using fundus images collected from a variety of sources.
First, a new multi-label retinal disease dataset, the MuReD dataset, is constructed, using a number of publicly available datasets for fundus disease classification. Next, a sequence of post-processing steps is applied to ensure the quality of the image data and the range of diseases, present in the dataset. 
For the first time in fundus multi-label disease classification, a transformer-based model optimized through extensive experimentation is used for image analysis and decision making.
Numerous experiments are performed to optimize the configuration of the proposed system. It is shown that the approach performs better than state-of-the-art works on the same task by 7.9\% and 8.1\% in terms of AUC score for disease detection and disease classification, respectively. The obtained results further support the potential applications of transformer-based architectures in the medical imaging field.

\end{abstract}

\begin{IEEEkeywords}
multi-label; fundus imaging; disease classification; transformer; deep learning
\end{IEEEkeywords}}

\maketitle

\IEEEdisplaynontitleabstractindextext

%
\IEEEpeerreviewmaketitle

\section{Introduction}
The retina is one of the main components of the eye, which supports the visual function. It is located at the back of the eye and its main job is to transform the light that enters through the eye to electrical signals that are passed on to the brain through the optical nerve. Due to its nature, the retina can both manifest the occurrence of diseases limited to the eyes, as well as broader scope physiological conditions, specifically, circulatory and brain diseases \cite{Mittal_Rajam_2020}.
    
    Diseases such as age-related macular degeneration (ARMD), diabetic retinopathy (DR), and glaucoma cause blindness to more than 10 million people around the world every year \cite{Mittal_Rajam_2020}. Indeed, glaucoma is the second most common cause of blindness in the developed world \cite{Badar_Haris_Fatima_2020}, with ARMD being the most common cause of blindness for people above 50 years old \cite{Badar_Haris_Fatima_2020}, and DR is one of the most important causes of vision loss for people in the age group from 25 to 74 years \cite{Mittal_Rajam_2020}.
    
    Regular examination of the retina may support the early diagnosis of diseases before the occurrence of any symptoms. Early diagnosis is crucial since early detection may prevent total vision loss in patients and support delaying and potentially stopping degenerative diseases, e.g., progressive retinal atrophy, through a timely treatment regime. 
    
    Automated analysis and diagnosis systems have a significant impact in medicine and biology \cite{han2011data}. Computer-aided analysis (CAD) of retinal images can, for instance, help physicians in disease diagnosis, and early treatment planning, reduce the time taken to process large datasets and minimize variability in image interpretation \cite{Mittal_Rajam_2020}. Moreover, automatic analysis offers several advantages over manual inspection, being more cost-effective, objective, reliable, and relaxing the requirement for trained specialists to grade images \cite{abramoff2010retinal}. On the other hand, manual inspection tends to be mundane, time-consuming, and requires proficient skills \cite{fraz2012blood}. Indeed, one of the major stumbling blocks for manual retinal examination in developing countries is the lack of a sufficient number of qualified medical personnel per capita to diagnose diseases \cite{stolte2020survey}.
    Prior to the development of deep learning methods, CAD systems were applied in various stages of the retinal diagnostic procedure, including image enhancement and restoration. Some CAD approaches attempted to imitate the means that clinicians perform retinal disease diagnosis, for instance, by performing image segmentation, feature extraction, and finally using machine learning \cite{6464593}. For example, in \cite{8102591}, a total of 32 local binary patterns (LBP) multi-scale texture features per image were used as an input to various classifiers, including instance-based multi-label learning model, Multi-label Support Vector Machine Learning, Multi-label Learning neural network Radial Basis Function and Back-Propagation Multi-label Learning with promising performance. In summary, state-of-the-art attempts at tackling multi-label retinal diagnosis relied on the use of feature engineering, coupled with traditional machine learning algorithms, which, however, have substantial limitations, in regards to the suitability and distinguishability of features in the context of multiple simultaneous retinal diseases.

    One of the most successful approaches to the automatic detection of retinal diseases is the use of Deep Learning (DL) techniques, specifically, Convolutional Neural Networks (CNN) and more recently, Transformer architectures. A considerable amount of work has been carried out on detecting the presence of common retinal diseases such as ARMD, DR, Glaucoma, etc. For instance, Zago et al., \cite{103537} developed a system that uses two CNNs (pre-trained VGG16 and CNN) to diagnose DR according to the probability of lesion patches. Jiang et al., \cite{8857160} developed a system based on three CNNs, namely, Inception-v3, ResNet152, and Inception-ResNet-v2, to classify fundus images into referable DR or non-referable DR. Burlina et al., \cite{103537} applied deep learning in detection and classification of ARMD using two deep convolutional networks, one was trained for the detection of ARMD, while the other used transfer learning. Finally, a diagnostic tool based on deep learning was developed for screening patients for common retinal diseases \cite{101016}.
    
    Despite the promising results obtained in the detection the occurrence of a single disease, the associated models are not sufficiently flexible to accommodate the simultaneous presence of multiple retinal diseases, which is commonly the case in real-world applications. Instead, clinicians require a diagnostic tool capable of detecting a diversity of conditions, simultaneously affecting a patient to provide the best possible treatment regime. Therefore, contrary to the vast majority of state-of-the-art works, which focus on detecting a single retinal disease, a higher value solution is multi-label disease classification, which would support simultaneous detection of a wide range of conditions present in a patient.
    
    There are various challenges when dealing with multi-label classification in fundus imaging. 
    For example, there is variability in the spatial extent of diseases, where some localize in specific regions of the retina, e.g., glaucoma, whereas others manifest themselves all over the retina (e.g., tessellation (TSLN) \cite{tsln}). Simultaneous diagnosis of such diseases thus requires powerful architectures, capable of detecting changes in both small and large spatial regions of the retina.
    
    Another issue is the scarcity of data. Most of the existing datasets are affected by different problems that make them unsuitable to satisfactorily train a multi-label model. For instance, they may focus on single diseases, contain a small number of samples, or indeed, a small number of pathologies to predict. Solving this problem requires the combination of existing datasets, to create an appropriate image database that can be used in model development.
    
    Finally, a common problem when working with multi-label datasets is class imbalance. Usually, a small number of disease labels contain most samples, whereas most labels have only a few images. Thus, techniques designed to deal with the class imbalance problem are required, either by suitably modifying the dataset or the means that the model learns from the data.
    
    In this work, a novel pipeline for multi-label disease classification based on fundus images is proposed.
    
    A new dataset that combines publicly available fundus datasets for both single and multi-disease detection is generated to address the scarcity of publicly available data and to alleviate the high-class imbalance present in publicly available datasets.
    Preprocessing is applied to preserve the quality of the data and generate a final retinal image dataset that contains a wide variety of diseases to predict with a sufficient number of samples per disease label.
    
    Finally, a transformer-based model optimized through extensive experimentation is employed for multiple retinal disease classification, using the proposed dataset. The model is trained using a novel scheme, optimized through a series of experiments on its architectural design and hyperparameters, and by using a variety of techniques to partly alleviate the effect of the class imbalance present in the MuReD dataset over the model performance.
    
    The main contributions of this work can be summarized as follows:
    \begin{enumerate}
        \item A new customized multi-label dataset, the MuReD dataset, is generated, which contains 20 disease classes, gathered from state-of-the-art sources and cleaned using an automatic quality score based on the sharpness and brightness of the image.
        
        \item A transformer-based model optimized through extensive experimentation is used for the first time to detect and classify multiple retinal diseases.
    \end{enumerate}

The remainder of this article is organized as follows. In Section \ref{sec:related_work}, an overview of existing techniques for multi-label classification using fundus imaging is given, together with a review of common methods to alleviate the class imbalance problem and the publicly available fundus datasets for retinal disease classification. In Section \ref{sec:dataset}, the steps to generate the proposed MuReD dataset are described in detail. Section \ref{sec:methods} describes the development of the transformer-based model. Section \ref{sec:res} presents the experiments carried out for the model training, and the comparison with state-of-the-art techniques. Finally, Section \ref{sec:conclusion} summarizes the main contributions of the research and proposes new avenues for future research.

\section{Related Work}
\label{sec:related_work}
    \subsection{CNN methods}
    A variety of works proposed the CNN architecture and its variants for multi-label disease classification on both public and private datasets. Cen et al., \cite{nature39} developed a deep learning platform (DLP) for the detection of 39 fundus diseases and conditions. For training, they used a combination of private datasets, collected from different regions of China, and the publicly available EyePACS dataset \cite{kaggle_2015}, reporting an AUC score of 0.99. Ju et al., \cite{longtailed} used a hybrid distillation approach to train a ResNet-50 \cite{resnet50}, extracting knowledge from two teachers, each trained with different sampling strategies. This approach used two private datasets, containing 100K and 1 million images, respectively, to classify 50 types of diseases. Finally, they reported a mean Average Precision (mAP) score of 64.14\% and 64.69\% for the 100K and 1 million image datasets, respectively.
    
    In terms of publicly available datasets, one of the most commonly used in multi-label disease classification is the ODIR dataset \cite{ODIR}. This dataset contains images of both eyes from 5,000 patients, each one classified into 8 different labels (1 for normal condition and 7 diseases). He et al., \cite{ODIR_attention} used a ResNet101 \cite{resnet50} model and a special attention module to find correlations between both images, reporting an AUC of 93\%. Li et al.,  \cite{ODIR_correlation} used a ResNet101 with a trainable Spatial Correlation Module (SCM) to find similarities between both images, reporting a similar AUC of 93\%.  Gour et al., \cite{odir_vgg16} trained a multi-input VGG16 architecture \cite{vgg16}, whereas Wang et al., \cite{odir_effnetb3} proposed an ensemble of two EfficientNets B3 \cite{efficientnet}, obtaining an AUC score of 84.93\% and 74\%, respectively.
    
    \subsection{Class Imbalance}
    Class imbalance is frequently encountered in multi-label datasets, as it is usual that a minority of classes contain the majority of data, whereas the majority of classes have a small number of samples in comparison. This is known as the long-tail distribution problem. 
    
    A literature review on techniques to deal with class imbalance, with a focus on multi-label problems was performed. There are four main approaches \cite{survey_multilabel} to address the class imbalance, with their effectiveness being dependent on the particular application:
    
\textbf{Resampling Methods:} \cite{ml_ros, lp_rus_ros, mlsmote} These methods are based on the pre-processing of the multi-label dataset, making it classifier independent. They can be divided into oversampling techniques, which generate new samples from the minority classes, and undersampling methods, which remove samples from the majority classes. Another categorization involves grouping into random methods and heuristic methods. Because of their model-independence property, this group of techniques is one of the most popular.

\textbf{Classifier Adaptation:} \cite{8672066, SUN2017375, he_gu_liu_2012} This requires the model to be designed to deal with the imbalance of the dataset. Although this technique can achieve competitive results, it is less popular because it requires expertise in both the classifier and the problem domain, and usually creates more complex and specialized training pipelines.

\textbf{Ensemble Approaches:} \cite{TAHIR20123738, TAHIR2012513, ense} This group of methods uses two or more models, each learning a different set of labels, and finally, the predictions of each model are combined to complete the full set of labels. The main disadvantage of this approach is that it requires substantial time and resources for training.

\textbf{Cost Sensitive Methods:} \cite{lin2018focal, asl} These methods employ custom metrics for the loss function, designed to increase the cost of misclassifying the minority classes, thus compensating for the difference of the samples in the majority classes. One of the most popular approaches is using weighted loss functions.
    
    When considering the aforementioned class imbalance methodologies, the use of ensemble methods was disregarded due to a large number of classes to predict in the multi-label dataset, i.e., 20 classes, see Section \ref{sec:dataset}), since this would require the training of several models, thus increasing the complexity of the overall system. Moreover, classifier adaptation methods were considered impractical due to the complexity of the selected model (see Section \ref{sec:methods}). Thus, both resampling and cost-sensitive methods were adopted to tackle this problem.
    
    There are various resampling algorithms in the multi-label setting, based on either random or heuristic resampling. 
    
    In Charte et al., \cite{lp_rus_ros} two random resampling algorithms, called LP ROS and LP RUS, for oversampling and undersampling, respectively, were proposed. These were based on the concept of Label Powerset transformation \cite{label_powerset}. These algorithms take the set of labels and generate a unique class for each unique combination, transforming efficiently a multi-label dataset into a multi-class dataset. After this procedure, the mean amount of positive samples per class is calculated and either the majority classes are lowered by dropping random samples, i.e., undersampling, or the minority classes are expanded by copying random samples, i.e., oversampling, to the mean value.
    
    Along a similar line of research, \cite{ml_ros} proposed two random resampling algorithms, called ML ROS and ML RUS, for oversampling and undersampling, respectively.
    These techniques make use of specific metrics to calculate the imbalance rate of a class and the mean imbalance rate of the entire dataset. For oversampling, if a class has an imbalance rate greater than the mean, random images from that class are copied until the mean is reached. A similar process is followed for the undersampling case.
    
    In the case of cost-sensitive methods, there are popular loss functions proposed for the problem of class imbalance, such as Weighted Binary Cross-Entropy (WBCE), Focal Loss \cite{lin2018focal} and more recently proposed loss functions for imbalanced datasets such as Assymetric Loss \cite{benbaruch2021asymmetric} and Polynomial Loss \cite{polyloss}, which achieved promising results in a variety of multi-label datasets.

    \subsection{Fundus image datasets}
    An extensive literature review to identify publicly available fundus image datasets for use in multi-label retinal disease classification was performed. There exists a variety of datasets in the literature, each one developed for a different task. Table \ref{tab:datasets} shows the available datasets. 
    
    The DRIVE dataset is one of the outputs of a diabetic retinopathy screening program in the Netherlands, consisting of 400 diabetic subjects between 25-90 years old. The STARE (STructured Analysis of the REtina) project was conceived and initiated in 1975, at the University of California, San Diego. In contrast to the DRIVE dataset, STARE contains several images demonstrating retinal abnormalities, thus exhibiting more variation. The CHASE-DB dataset was developed during the Child Heart Health Study in England (CHASE), a cardiovascular health survey in 200 primary schools in London, Birmingham, and Leicester. It captures information from 19 pupils from 10 primary schools, who were measured both in the morning and the afternoon on the same day between September 2007 and March 2008 by the same observer.
    The Messidor dataset contains 1200 eye fundus color numerical images acquired by three ophthalmologic departments using the same color video camera. Two diagnoses were provided by medical experts: retinopathy grade (4 grades) and risk of macular edema.
    The e-ophtha dataset was designed for scientific research in Diabetic Retinopathy. It is composed of two databases, one containing 47 images with exudates and 35 images with no lesion, and the other one containing 148 images with microaneurysms and 233 images with no lesion.
    The Kaggle-EyePacs dataset contains high-resolution retina images provided by the EyePacs platform. These images were rated by a clinician for the presence of diabetic retinopathy with 5 different grade levels. 
    The ARIA (Automated Retinal Image Analysis) dataset was collected in the United Kingdom between 2004 and 2006 from adult males and females. The images come from three control groups: healthy, age-Related macular degeneration (ARMD), and diabetic patients.
    The RFMiD (Retinal Fundus Multi-disease Image Dataset) consists of 3200 fundus images captured by three different cameras. It contains 46 pathologies that appear in routine clinical settings annotated by consensus from two senior retinal experts.
    
    \begin{table}[h!]
    \footnotesize
  \centering
  \setlength\tabcolsep{3pt}
\begin{tabular}{@{}rcl@{}}
\toprule
\multicolumn{1}{c}{Dataset} & No. images & \multicolumn{1}{c}{Tasks}                                                                                           \\ \midrule
\rowcolor[HTML]{EFEFEF} 
DRIVE \cite{niemeijer2004drive} & 400        & Blood Vessel Demarcation                                                                                              \\
STARE \cite{stare} & 388        & \begin{tabular}[c]{@{}l@{}}Blood Vessel Demarcation\\ Multi-label Classification\end{tabular}                               \\
\rowcolor[HTML]{EFEFEF} 
CHASE\_DB1 \cite{6224174} & 28         & Blood Vessel Demarcation                                                                                              \\
Messidor \cite{decenciere2014feedback} & 1200 & \begin{tabular}[c]{@{}l@{}}DR grading\\ Macula Edema Risk grade\end{tabular} \\
\rowcolor[HTML]{EFEFEF} 
E-Ophtha \cite{decenciere2013teleophta}  & 463       & \begin{tabular}[c]{@{}l@{}}Exudates Demarcation\\ Mycroaneurisms Demarcation\end{tabular} \\ 
Kaggle-EyePacs \cite{kaggle_2015}  & 88702       & DR grading \\
\rowcolor[HTML]{EFEFEF} 
ARIA \cite{aria} & 143        & \begin{tabular}[c]{@{}l@{}}Blood Vessel Demarcation\\ Optic disc and Fovea location\\ ARMD and DR labels\end{tabular} \\
RFMiD  \cite{RFMiD} & 3200       & Multi-label classification                                                                                                 \\ \bottomrule
\end{tabular}
\caption{Publicly available fundus image datasets}
\label{tab:datasets}
\end{table}

From the identified datasets, some shared problems were observed, specifically on those designed for multi-label tasks. The first problem is the low number of samples present for the underrepresented diseases, where half of them contain a maximum of 20 images and as few as one image, which significantly reduces the confidence of any model in classifying these diseases. The second problem is the high-class imbalance present in the identified datasets, where all of them show a long-tail distribution problem with substantial differences in samples between the overrepresented and underrepresented disease labels. Finally, the third problem is the lack of guarantee about the image's quality, since all of the multi-label datasets do not perform any cleaning steps or ensure any degree of quality in the images. An example of these low-quality images can be appreciated in Figure \ref{fig:bm-images}
    
\section{Dataset}
\label{sec:dataset}

To address the limitations in the publicly available datasets, a new custom dataset was constructed, i.e., the MuReD (\textbf{Mu}lti-label \textbf{Re}tinal \textbf{D}iseases) dataset. The purpose of this new dataset is to have:
\begin{enumerate}
    \item A sufficiently large number of eye diseases with a sufficient number of samples per disease class.
    
    \item A certain degree of quality in the images contained in the dataset.
\end{enumerate}

For the first point, the ideal dataset contains a wide variety of diseases to classify, with sufficient samples per disease to learn it effectively, and, at the same time, does not present a high degree of class imbalance.

For the second point, ensuring a certain degree of quality in fundus images is crucial since they tend to show significant variations in quality caused by both pathogenic factors, i.e., cataracts, or external factors, i.e., equipment misuse, environmental conditions, poor training, etc. \cite{fundus_quality}. All these factors degrade the quality of the fundus image by inserting noise, blurriness, and artifacts which increase uncertainty and the risk of misclassification.

The MuReD dataset was constructed using some of the publicly available datasets, to have a wide variety of diseases to predict, from a variety of sources, with varying image quality and at the same time ensuring a minimal degree of quality, to make the model more robust against image variations and a sufficient number of samples per disease class, so that the model can learn them effectively.
 
 The MuReD dataset is composed of the ARIA dataset \cite{aria}, containing 143 images and three labels to predict, the STARE dataset \cite{stare}, with 388 images and 21 conditions, and the training set of the RFMiD dataset \cite{RFMiD}, which consists of 1920 images with 46 different pathologies. It was decided to incorporate only the training set of the RFMiD dataset into the MuReD dataset to avoid too much bias since both ARIA and STARE datasets contain a smaller number of images in comparison to the full RFMiD dataset.
 Thus, the first version of the new composite dataset consisted of 2451 samples, 52 disease labels, one "NORMAL" class for healthy fundus images, and the "OTHER" class that is used to indicate the presence of a rare disease from which very few samples are available to consider it a class by its own.

\subsection{Dataset Cleaning}
Several cleaning procedures were performed on the first version of the composite dataset to eliminate labels with a small number of samples while ensuring that the overall quality of the images was sufficient for model development purposes.

First, it was observed that several labels in the original dataset, contained a low number of samples, and thus, they would not benefit from data augmentation techniques, while model performance would also be affected. Experiments were conducted on the percentage of modified samples and labels to identify the optimal threshold, i.e., the minimum number of images per label, to consider whether a label should be included or not. The "usable" labels would be kept in the dataset, whereas the "not usable" ones would be dropped, and the samples that were part of these labels would be included in the "OTHER" class. 
Following an extensive number of simulation studies, it was concluded that the best threshold to use was 30 samples, thus, ending up with a total of 20 classes for prediction purposes, including the "OTHER" class, which accounts for 10

Next, it was noticed that there are several instances, where no information could be acquired due to poor lighting conditions, such as high brightness or complete black zones, excessive blurring, etc. To detect low-quality images, an image quality score was calculated, based on the blur metric proposed by Kanjar and Masilamani \cite{blur_measure}, which measures the sharpness and brightness of an image, using edge detection and neighboring pixel difference information.

To measure image quality, the edges present in the image had to be detected first. In the original version of the work, the use of the Sobel operator was suggested, however, through empirical observation and experimentation, it was concluded that the Canny edge detector performs better on the given image dataset. Consider an image $I$, the extracted set of edges $E$, and $N_{xy}$ as the set of 8-neighbors of the pixel $I(x, y)$, $I(x, y) \in E$, then the blur metric is given by:

\[BM = \frac{\sum_{I(x,y)\in E}\sqrt{\sum_{I(x',y')\in N_{xy}}\frac{\{I(x,y)-I(x',y')\}^2}{|N_{xy}|}}}{\sum_{I(x,y)\in E}I(x,y)}\]

where $|N_{xy}|$ represents the cardinality of the set $N_{xy}$.

The concept behind the use of this metric is that good quality images have high levels of sharpness and a low amount of blur, and thus, for a sharp image, the intensity changes near edges will be significant, whereas, in the case of a blurred instance, they will be low. A higher value in the blur metric translates to the image having higher sharpness, whereas a lower value means the amount of blurring is high.

To evaluate the suitability of this method and the associated results, 150 images were visually identified and selected which contained the lowest perceived image quality and compared to the 150 images with the worst score, automatically determined by the blur metric. For instance, 90\% of the images identified manually were identical to the ones with the lowest blur metric.

\begin{figure*}[h!]
\centering
\includegraphics[scale=0.7]{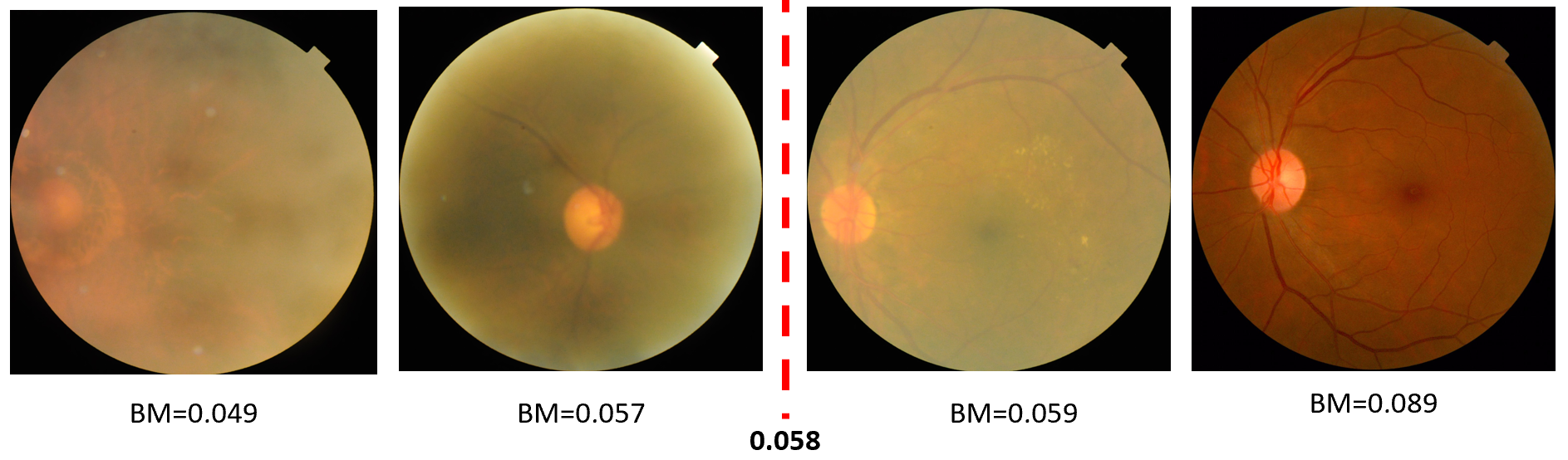}
\caption{Examples of images with their blur measures. A blur metric threshold of 0.058 was used. Images below the threshold were dropped (left of the red line), whereas images above the threshold were used in the dataset (right of the red line).}
  \label{fig:bm-images}
\end{figure*}

Next, the blur metric score was employed to sort the images from the highest to the lowest score, and during this ranking process, a quality threshold was determined, i.e., a score value corresponding to images of acceptable quality. Following the empirical observation, it was decided to drop the bottom 10\% of the images, using a blur score threshold of 0.058, since most of the images below this value are excessively blurred or contain artifacts that substantially impact the image quality. Figure \ref{fig:bm-images} shows examples of images, which were dropped as they were below the threshold, and images included in the dataset.

\begin{figure*}[t]
\centering
\includegraphics[scale=1.0]{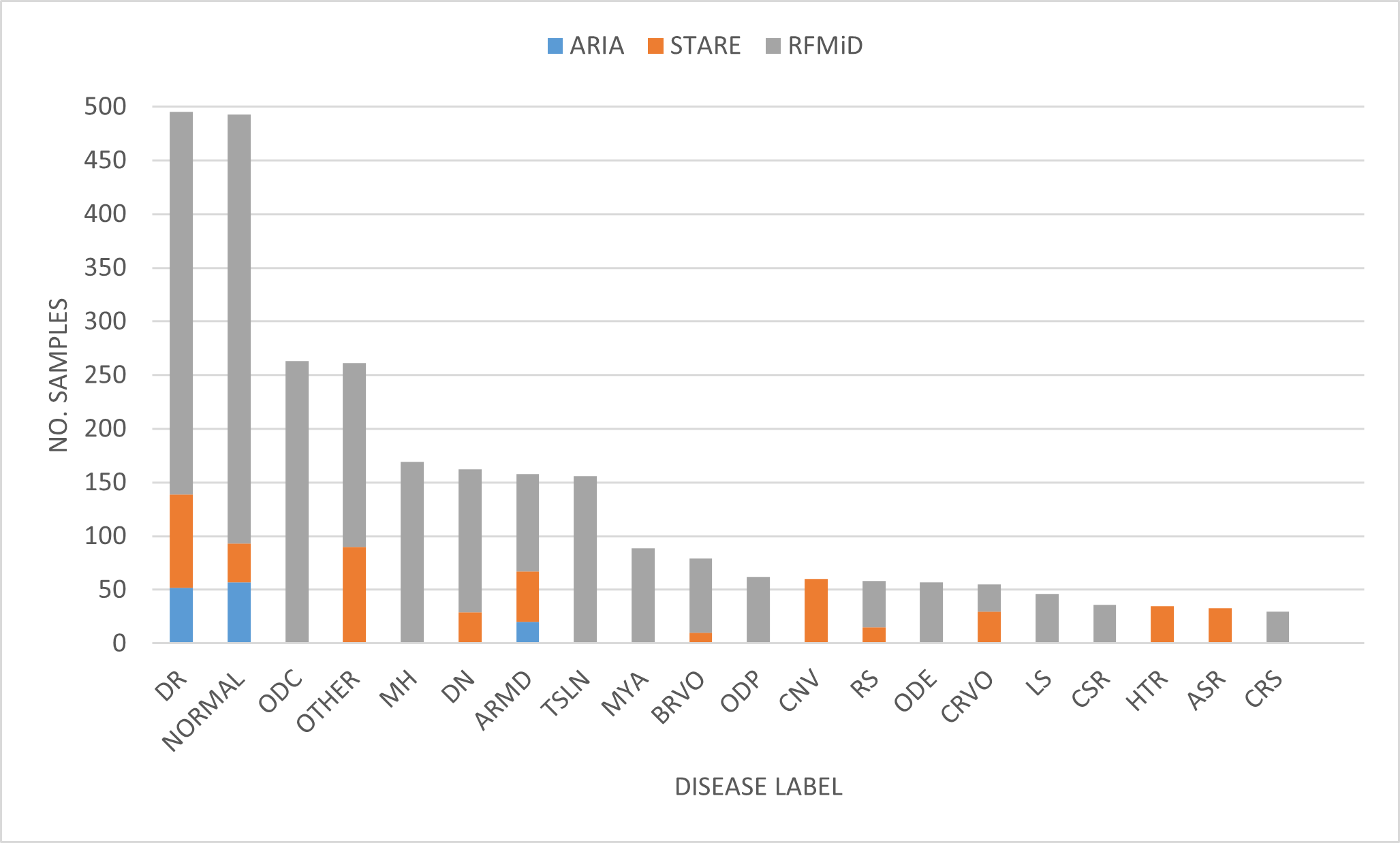}
\caption{Overview of dataset distribution. The plot shows the amount of samples per label and the proportion of the contribution of the three original datasets.}
  \label{fig:dataset}
\end{figure*}

Following this cleaning phase, the final dataset consisted of 2208 samples with 20 disease labels. The label distribution of the fine-tuned dataset is shown in Figure \ref{fig:dataset}. Details of the classes and the numbers of samples per label are given in Table \ref{tab:dataset_info}.

\subsection{Comparison with publicly available datasets}
To illustrate and quantify the improvement on the class imbalance problem that was achieved by the creation of the MuReD dataset, comparisons were performed with the available datasets, i.e., ARIA, STARE, and RFMiD.

To perform this comparison, two metrics proposed by Charte et al., \cite{ml_ros} were used, designed specifically to measure the imbalance present in multi-label datasets. The first metric, i.e., the mean Imbalance Rate (meanIR), measures the mean ratio of samples present in any label compared with the label that contains the majority of samples. The second metric, i.e.,  the Coefficient of Variation of Imbalance Rate per Label (CVIR), indicates if all the labels suffer from a similar level of imbalance or, on the other hand, there are large differences in them. The higher the CVIR, the higher this difference.

Using the proposed evaluation metrics, it is demonstrated that the MuReD dataset can reduce the meanIR by 5.59, i.e., 44\% reduction, and the CVIR by 0.21, i.e., 23\% reduction, compared with the best values achieved by the available datasets.

\begin{table}[h!]
\footnotesize
  \centering
  \setlength\tabcolsep{3pt}
\begin{tabular}{@{}clccc@{}}
\toprule
Acronym & \multicolumn{1}{c}{Full Name}  & Training & Validation & Total \\ \midrule
\rowcolor[HTML]{EFEFEF}DR      & Diabetic Retinopathy           & 396      & 99         & 495   \\
NORMAL  & Normal Retina                  & 395      & 98         & 493   \\
\rowcolor[HTML]{EFEFEF}MH      & Media Haze                     & 135      & 34         & 169   \\
ODC     & Optic Disc Cupping             & 211      & 52         & 263   \\
\rowcolor[HTML]{EFEFEF}TSLN    & Tessellation                    & 125      & 31         & 156   \\
ARMD    & Age-Related Macular Degeneration    & 126      & 32         & 158   \\
\rowcolor[HTML]{EFEFEF}DN      & Drusen                         & 130      & 32         & 162   \\
MYA     & Myopia                         & 71       & 18         & 89    \\
\rowcolor[HTML]{EFEFEF}BRVO    & Branch Retinal Vein Occlusion  & 63       & 16         & 79    \\
ODP     & Optic Disc Pallor              & 50       & 12         & 62    \\
\rowcolor[HTML]{EFEFEF}CRVO    & Central Retinal Vein Oclussion & 44       & 11         & 55    \\
CNV     & Choroidal Neovascularization   & 48       & 12         & 60    \\
\rowcolor[HTML]{EFEFEF}RS      & Retinitis                      & 47       & 11         & 58    \\
ODE     & Optic Disc Edema               & 46       & 11         & 57    \\
\rowcolor[HTML]{EFEFEF}LS      & Laser Scars                    & 37       & 9          & 46    \\
CSR     & Central Serous Retinopathy     & 29       & 7          & 36    \\
\rowcolor[HTML]{EFEFEF}HTR     & Hypertensive Retinopathy       & 28       & 7          & 35    \\
ASR     & Arteriosclerotic Retinopathy   & 26       & 7          & 33    \\
\rowcolor[HTML]{EFEFEF}CRS     & Chorioretinitis                & 24       & 6          & 30    \\
OTHER   & Other Diseases                 & 209      & 52         & 261   \\ \bottomrule
\end{tabular}
\caption{Class labels and number of samples per label in the MuReD dataset, following the application of the cleaning procedure.}
\label{tab:dataset_info}
\end{table}

\section{Methods}
\label{sec:methods}

\subsection{Data Preprocessing}
Visual examination of the images in the dataset revealed that most of the retinal information is at the center of the image, surrounded by a black background, which does not contain useful information and can affect the performance and training time of the model, due to the presence of redundant information and larger image size.

Thus, background removal, also known as field-of-view (FOV) extraction, was performed using the method proposed by Kulkarni et al., \cite{fov_extraction}. This works by taking advantage of the sudden changes in brightness between the dark background and the region of interest (ROI), i.e., the part of the image that contains the retina. Specifically, two centerline scans (horizontal and vertical) are performed over the red channel of the image and a threshold is set to \(th = max(I) \times 0.06\), where $I$ represents the intensity values of each scan line. The value of $0.06$ was proposed by the authors after extensive empirical testing. 

\subsection{Multi-Label Classification Model}
The C-Tran architecture, proposed by Lanchantin et al., \cite{c_tran}, was selected as the classification model. The model was specifically designed for multi-label tasks, demonstrating high-performance rates on popular multi-label datasets, such as MS-COCO \cite{ms-coco} in its multi-label version of 80 categories, and Visual Genome \cite{vg500} using the most common 500 categories. 

\begin{figure*}[h!]
\centering
\includegraphics[scale=0.7]{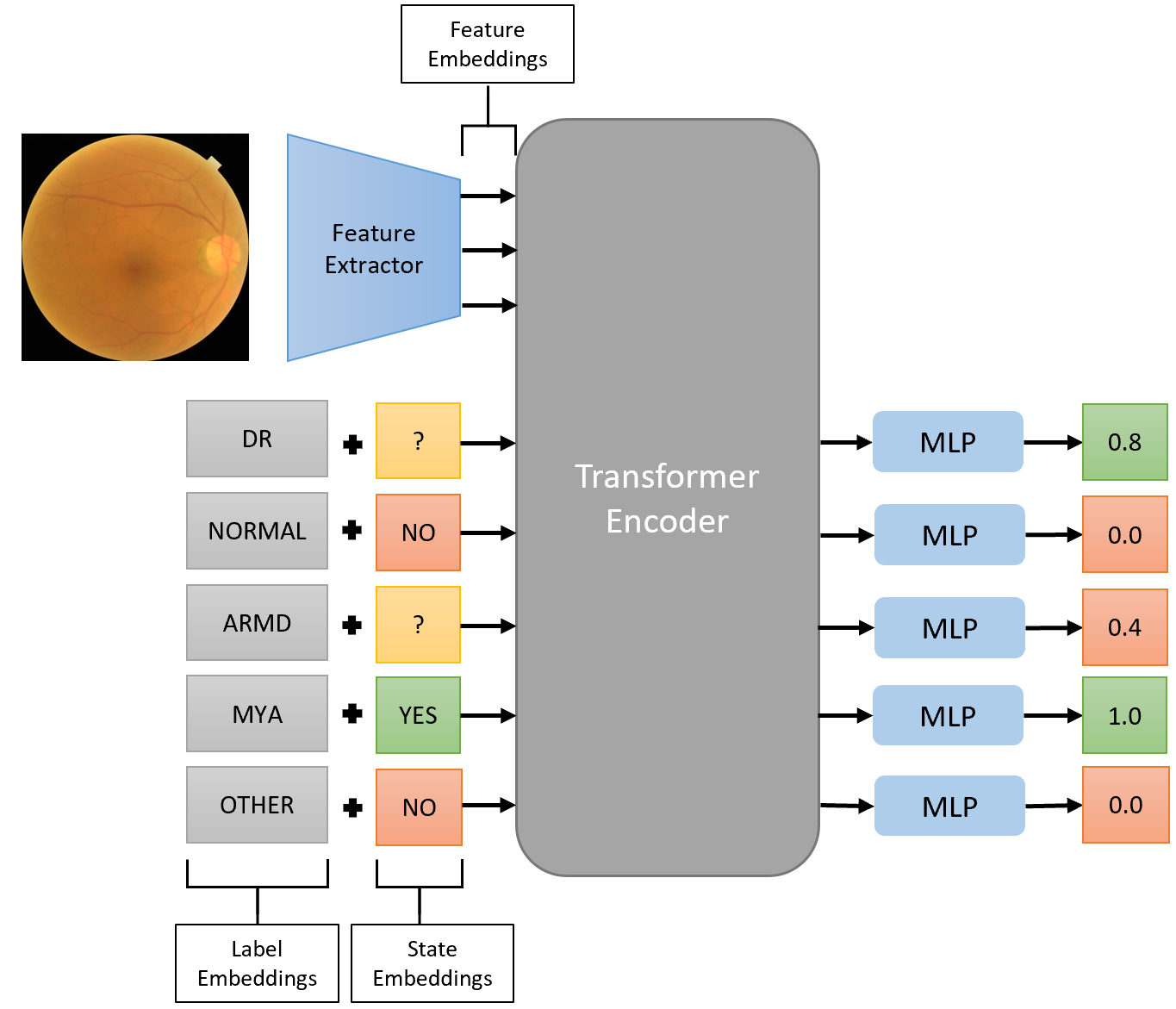}
\caption{C-Tran architecture. A feature extractor is used to generate the set of feature embeddings. Then, the label embeddings are combined with the state embeddings to train the model using partial information. Finally, the output from the transformer is used to feed the MLP head to output the probabilities for the unknown classes.}
  \label{fig:c-tran}
\end{figure*}

The C-Tran model consists of a Transformer encoder that feeds from both visual features extracted by a CNN and a set of masked labels. This formulation is possible due to the order invariant characteristic of transformers, which allows any type of dependency between all features and labels to be learned.
A general overview of the C-Tran architecture is shown in Figure \ref{fig:c-tran}

The C-Tran architecture is composed of 3 main parts.
In the first part, a set of features, labels, and state embeddings is generated, serving as the inputs to the transformer encoder. 
For an input image, $x$, a set of visual features $Z = (h \times w \times d) $ are extracted with the use of the CNN backbone. Then, a set of patches $P = (h \times w)$ can be generated from each dimension $d$ from the original set $Z$. These patches are used as input to the transformer encoder.  

For each image, a set of label embeddings $L=\{l_1,l_2,...,l_l\}$ is generated, each $l_i$ of size $d$, which represent the \(l\) different labels in the ground truth $y$ via a learned embedding layer of size \(d \times l\).

With the use of these label embeddings, it is straightforward to add knowledge using a state embedding vector $s_i$ of size $d$:
\[\tilde{l_i} = l_i + s_i\]
Using the state vector $s_i$, Three states can be represented: Unknown (U) with a value of 0, Negative (N) with a value of -1, and Positive (P) with a value of +1.
State embeddings add significant value to the model training by using partially labeled data, extra labels, or no prior knowledge.

The second part focuses on modeling the feature and label interactions. A  transformer encoder is used as the model because of its ability to capture dependency information between variables. Also, its order invariant characteristics make it suitable to find dependencies between features and labels.
Given the set of embeddings $H = \{z_1,...,z_{h \times w}, \tilde{l}_1,...,\tilde{l}_l \}$, the weight between $h_i$ and $h_j$, represented as $\alpha_{ij}$, is calculated using the self-attention mechanism. 
First, the normalized scalar attention coefficient $\alpha_{ij}$ is computed for all embedding pairs $i$ and $j$. Then, the coefficient $\alpha_{ij}$ is used to update $h_i$ to $h'_i$ with a weighted sum of all the embeddings. A non-linear ReLU is applied at the end of this process. The formulas for this process are presented as follows:

\[\alpha_{ij} = softmax((W^qh_i)^\top(W^kh_j)/\sqrt{d})\]
\[\bar{h}_i = \sum_{j=1}^M \alpha_{ij}W^vh_j\]
\[h'_i = ReLU(\bar{h}_iW^r + b_1)W^o + b_2\]

where $W^k$, $W^q$, $W^v$ are the key, query and value weight matrices, respectively, $W^r$ and $W^o$ are the transformation matrices, and $b_1$, $b_2$ are bias vectors. The output vector $H'=\{z'_1,...,z'_{h \times w},\tilde{l}'_1,...,\tilde{l}'_l\}$ can be used as the input to the next transformer encoder layers, and this update procedure is repeated.

The final label predictions are computed using independent feed-forward networks ($FFN_i$) for each label embedding $l'_i$ using a single linear layer of size $d$ and the sigmoid function.

The described architecture allows to easily add prior knowledge to the C-Tran using the state embeddings. However, to make it flexible enough to handle any amount of known labels during inference, the authors proposed a novel training scheme, label mask training (LMT), used to help the model both learn label correlations and perform inference with any number of known labels.

LMT masks a random number of labels, i.e., from 25\% to 100\%, for each sample by adding the unknown state to its label embeddings. The rest are set with the known state, i.e., from 0\% to 75\%, either positive or negative. The model then predicts the unknown labels, and the loss is calculated to update the model parameters. By masking random amounts of labels, the model can learn many possible known label combinations and handle any inference setting, e.g., regular, partial, or extra labels.

This approach yields better results than other techniques that exploit label relations such as Graph Convolutional Networks (GCN). In this work, the LMT approach for training and no prior knowledge for inference is used.

In this approach, the LMT scheme is employed for training to allow the model to better learn label correlations by using prior knowledge. However, the focus of this work is to perform regular predictions while inferring, i.e., classification without prior or partial knowledge. During inference, all the label embeddings are masked with the unknown state, effectively hiding all prior or partial information to the model by replacing it with all-zero embeddings.

\section{Experimental Setup and Results}
\label{sec:res}

\begin{table*}[h!]
\centering
\begin{tabular}{lllc}
\toprule
Augmentation             & \multicolumn{1}{c}{Description}                       & \multicolumn{1}{c}{Parameters}                                   & Probability \\ \midrule
\rowcolor[HTML]{EFEFEF}HorizontalFlip           & Flips the image horizontally                          & -                                                                & 0.5         \\
VerticalFlip             & Flips the image vertically                            & -                                                                & 0.5         \\
\rowcolor[HTML]{EFEFEF}Rotate                   & \begin{tabular}[c]{@{}l@{}}Randomly rotates the image by an angle \\ specified by the maximum angle limit\end{tabular}  & limit=30                                                         & 0.5         \\
MedianBlur               & Applies median filter to the image                    & blur\_limit=7                                                    & 0.3         \\
\rowcolor[HTML]{EFEFEF}GaussNoise               & Applies Gaussian noise to the image                   & var\_limit=(0.38)                                                & 0.5         \\
HueSaturationValue       & \begin{tabular}[c]{@{}l@{}}Randomly changes the hue, saturation and value \\ of the image \end{tabular} & hue\_shift\_limit=10, sat\_shift\_limit=10, val\_shift\_limit=10 & 0.3         \\
\rowcolor[HTML]{EFEFEF}RandomBrightnessContrast & \begin{tabular}[c]{@{}l@{}}Randomly changes the brightness and contrast \\ of the image \end{tabular} & brightness\_limit=(-0.2, 0.2), contrast\_limit=(-0.2, 0.2)       & 0.3         \\
Cutout                   & Randomly crops square regions on the image              & max\_h\_size=20, max\_w\_size=20, num\_cutout\_regions=5                   & 0.5         \\ \bottomrule
\end{tabular}
\caption{List of augmentations used during training.}
\label{tab:augmentations}
\end{table*}

\subsection{Metrics}
To evaluate the performance of the selected model, the scoring metric proposed in the RIADD challenge \cite{RIADD} was used, as it gives equal importance to the correct detection of the presence of disease and its correct classification. 
First, the F1, mAP, and AUC scores are calculated for all labels in the dataset. Then, the set $T$ is defined as the set of labels, representing a disease label. Using the scores from set $T$, the average score for each metric of the disease classes (only excluding the "NORMAL" label) is computed and named $ML\_mAP$, $ML\_F1$, and $ML\_AUC$ respectively.
These metrics are given by: 



\[AP = \sum_{i=0}^{|T|-1} [recall_i - recall_{i+1}] \times precision_i\]
\[ML\_mAP = \frac{1}{|T|}\sum_{i=1}^{|T|} AP_i\]

\[ML\_F1 = \frac{1}{|T|}\sum_{i=1}^{|T|} F1_i\]

\[ML\_AUC = \frac{1}{|T|}\sum_{i=1}^{|T|} AUC_i\]

The two most important metrics, used to evaluate the performance of the models, are the $ML\_Score$, calculated by the average of $ML\_mAP$ and $ml\_AUC$, and the $Model\_Score$, which is the average of the $ML\_Score$ and the AUC score of the "NORMAL" class, termed $Bin\_AUC$.

\[ML\_Score = \frac{ML\_mAP + ML\_AUC}{2}\]
\[Model\_Score = \frac{ML\_Score + Bin\_AUC}{2}\]

The last metric, $Bin\_F1$, represents the F1-score of the "NORMAL" label.

\subsection{Determination of optimal model configuration}
Three sets of experiments were performed to determine the optimal configuration of the C-Tran model on the MuReD dataset.

In the first set of experiments, both traditional and state-of-the-art CNN models were tested as backbones, i.e., feature extractors, for the C-Tran architecture to find the most suitable and best-performing for this task. Next, the second set of experiments focused on alleviating the effect of the class imbalance present in the MuReD dataset on the model performance. The approach employed to address this problem was to test different resampling techniques designed for multi-label datasets and to implement different loss functions, including some designed to alleviate class imbalance.  Finally, the third set of experiments focuses on finding the most suitable values for certain hyperparameters, i.e., image size and batch size, so as to design the optimal model configuration.

For all experiments, the C-tran used the Adam optimizer \cite{kingma2017adam}, with a batch size of 16, BCE loss function, a learning rate (LR) of $10^{-5}$, three transformer encoder layers, image size of 384$\times384,$ and a dropout rate of 0.1. Each time a new best-performing value or method for any hyperparameter is found, it replaces the one proposed in this base configuration.

Finally, to increase the amount and variety of samples presented to the model during training on each batch, different random augmentations were used, based on the configuration proposed by the RIADD challenge winner \cite{hanson0910}. The Python's albumentations library \cite{info11020125} was used to generate the set of augmentations. Table \ref{tab:augmentations} shows a detailed description of the used augmentations.

\subsubsection{Backbone Selection}
For testing different CNN backbones, first traditional and well-known architectures such as InceptionV3 \cite{inceptionv3} and VGG16 \cite{vgg16} were tested as a baseline. Next, both EfficientNet \cite{pmlr-v97-tan19a} and EfficientNetV2 \cite{tan2021efficientnetv2} were considered since they have become the go-to architecture for most of the vision tasks due to their excellent performance and small size. After that, the ResNet101 \cite{resnet50} architecture was tested, including one of its most popular variations, Wide ResNet101 \cite{wide_resnet}. It was decided to include ResNext architectures \cite{8100117} given that it is one of the top performers in vision tasks. Finally, the DenseNet161 \cite{densenet} architecture was added to the experiments due to its competitive performance on ImageNet.

For the EfficientNet models, the pre-trained versions using the noisy student training technique \cite{9156610} were used, since they achieved a better score on ImageNet. For EfficientNetV2, the models pre-trained on ImageNet-21K were used. For ResNext\_32x4d, the model pre-trained with the semi-weakly supervised learning technique proposed by \cite{yalniz2019billionscale} was used, since it achieved better results than using conventional training. For the rest of the architectures, i.e., InceptionV3, VGG16, ResNet101, WideResNet101, DenseNet161 and ResNext\_32x8d, the pre-trained version provided by the Torchvision package \cite{torchvision} was used. Table \ref{tab:backbone_comparison} shows the results of this experiment.

\begin{table}[h!]
\footnotesize
  \centering
  \setlength\tabcolsep{3pt}
\begin{tabular}{@{}rccccccc@{}}
\toprule
\multicolumn{1}{c}{Backbone}         & \begin{tabular}[c]{@{}c@{}}ML\\ F1\end{tabular} & \begin{tabular}[c]{@{}c@{}}ML\\ mAP\end{tabular} & \begin{tabular}[c]{@{}c@{}}ML\\ AUC\end{tabular} & \textbf{\begin{tabular}[c]{@{}c@{}}ML\\ Score\end{tabular}} & \begin{tabular}[c]{@{}c@{}}Bin\\ AUC\end{tabular} & \begin{tabular}[c]{@{}c@{}}Bin\\ F1\end{tabular} & \textbf{\begin{tabular}[c]{@{}c@{}}Model\\ Score\end{tabular}} \\ \midrule
\rowcolor[HTML]{EFEFEF} 
{InceptionV3}   & {0.469}                    & {0.569}                     & { 0.933}                     & { 0.751}                                & { 0.951}                      & { 0.755}                     & {0.851}                                   \\
EfficientNetB5                       & 0.501                                           & 0.625                                            & 0.943                                            & 0.784                                                       & 0.965                                             & 0.825                                            & 0.874                                                          \\
\rowcolor[HTML]{EFEFEF} 
EfficientNetB6                       & 0.504                                           & 0.627                                            & 0.946                                            & 0.787                                                       & 0.964                                             & 0.789                                            & 0.875                                                          \\
{WideResNet101} & {0.537}                    & {0.638}                     & {0.945}                     & {0.791}                                & {0.960}                      & {0.794}                     & {0.876}                                   \\
\rowcolor[HTML]{EFEFEF} 
{VGG16}         & {0.508}                    & {0.622}                     & {0.940}                     & {0.781}                                & {\textbf{0.977}}             & {\textbf{0.837}}            & {0.879}                                   \\
EfficientNetV2-M                     & 0.570                                           & 0.683                                            & 0.955                                            & 0.819                                                       & 0.958                                             & 0.781                                            & 0.889                                                          \\
\rowcolor[HTML]{EFEFEF} 
EfficientNetV2-L                     & 0.585                                           & 0.680                                            & 0.954                                            & 0.817                                                       & 0.961                                             & 0.806                                            & 0.889                                                          \\
ResNext101 32x4d                     & 0.585                                           & 0.677                                            & 0.953                                            & 0.815                                                       & 0.964                                             & 0.802                                            & 0.889                                                          \\
\rowcolor[HTML]{EFEFEF} 
ResNext101 32x8d                     & \textbf{0.612}                                  & 0.683                                            & 0.947                                            & 0.815                                                       & 0.966                                             & 0.785                                            & 0.890                                                          \\
ResNet101                            & \textbf{0.612}                                  & \textbf{0.689}                                   & 0.955                                            & 0.822                                                       & 0.970                                             & 0.808                                            & 0.896                                                          \\
\rowcolor[HTML]{EFEFEF} 
{DenseNet161}   & {0.595}                    & {\textbf{0.689}}            & {\textbf{0.957}}            & {\textbf{0.823}}                       & {0.973}                      & {0.822}                     & {\textbf{0.898}}                          \\ \bottomrule
\end{tabular}
\caption{Comparison of results for different CNN architectures as feature extractors.}
\label{tab:backbone_comparison}
\end{table}

This comparison shows that DenseNet161 performs best in this task, compared to other traditional and state-of-the-art CNN architectures. Following experimentation makes use of DenseNet161 as the C-Tran backbone.

\subsubsection{Class Imbalance}
To determine a suitable method to reduce the effect of class imbalance on system performance, two experiments were performed, which reflected two popular approaches, i.e., resampling methods and weighted loss functions. In these experiments, the best-performing backbone from the previous experiment was used, i.e., DenseNet161, along with a
LR of $10^{-5}$, the Adam optimizer, and the BCE loss (only in the resampling experiment).

The first experiment focused on resampling algorithms, utilizing random oversampling and undersampling to determine which method would be more beneficial to the model. Oversampling was performed using the LP ROS and ML ROS algorithms, whereas undersampling was performed using the LP RUS and ML RUS techniques. A resampling percentage of 10\% was used for all methods to see which one would be the best-performing and to scale this value later for further improvement. Table \ref{tab:resampling_comparison} shows the results of the different resampling techniques.

\begin{table}[h!]
\footnotesize
  \centering
  \setlength\tabcolsep{3pt}
\begin{tabular}{@{}lccccccc@{}}
\toprule
\multicolumn{1}{c} {Algorithm} & 
\begin{tabular}{c}ML \\ F1 \end{tabular} & 
\begin{tabular}{c}ML \\ mAP \end{tabular} & 
\begin{tabular}{c}ML \\ AUC \end{tabular} & \begin{tabular}{c}\textbf{ML} \\ \textbf{Score} \end{tabular} & \begin{tabular}{c}Bin \\ AUC \end{tabular} & 
\begin{tabular}{c}Bin \\ F1 \end{tabular}  & \begin{tabular}{c}\textbf{Model} \\ \textbf{Score} \end{tabular} \\ \midrule
\rowcolor[HTML]{EFEFEF} 
{LP RUS 10\%}   & {0.544}          & {0.656}          & {0.959}          & {0.807}          & {0.960}          & {0.778}          & {0.884}          \\
\rowcolor[HTML]{FFFFFF} 
{ML RUS 10\%}   & {0.582}          & {0.676}          & {0.959}          & {0.817}          & {0.962}          & {0.774}          & {0.889}          \\
\rowcolor[HTML]{EFEFEF} 
{No Resampling} & {\textbf{0.595}} & {0.689}          & {0.957}          & {0.823}          & {\textbf{0.973}} & {\textbf{0.822}} & {0.898}          \\
\rowcolor[HTML]{FFFFFF} 
{ML ROS 10\%}   & {0.579}          & {\textbf{0.697}} & {0.959}          & {\textbf{0.828}} & {0.968}          & {0.759}          & {0.898}          \\
\rowcolor[HTML]{EFEFEF} 
{LP ROS 10\%}   & {0.585}          & {0.693}          & {\textbf{0.962}} & {0.827}          & {0.971}          & {0.778}          & {\textbf{0.899}} \\ \bottomrule

\end{tabular}
\caption{Comparison of results obtained by different resampling algorithms.}
\label{tab:resampling_comparison}
\end{table}

As it can be appreciated, the ML ROS algorithm maintained the same performance whereas the LP ROS algorithm achieved a slight increase. Since the performance difference between these two algorithms was tiny, it was decided to conduct further tests on both methods by optimizing their resampling percentage to evaluate whether any major improvement could be achieved. Table \ref{tab:ml_ros_comparison} shows the results of testing different resampling percentages for the ML ROS method. Table \ref{tab:lp_ros_comparison} shows the results of testing different resampling percentages for the LP ROS algorithm.

\begin{table}[h!]
\footnotesize
  \centering
  \setlength\tabcolsep{3pt}
\begin{tabular}{@{}lccccccc@{}}
\toprule
\multicolumn{1}{c} {Algorithm} & 
\begin{tabular}{c}ML \\ F1 \end{tabular} & 
\begin{tabular}{c}ML \\ mAP \end{tabular} & 
\begin{tabular}{c}ML \\ AUC \end{tabular} & \begin{tabular}{c}\textbf{ML} \\ \textbf{Score} \end{tabular} & \begin{tabular}{c}Bin \\ AUC \end{tabular} & 
\begin{tabular}{c}Bin \\ F1 \end{tabular}  & \begin{tabular}{c}\textbf{Model} \\ \textbf{Score} \end{tabular} \\ \midrule
\rowcolor[HTML]{EFEFEF} 
{No Resampling} & {0.595}          & {0.689}          & {0.957}          & {0.823}          & {\textbf{0.973}} & {0.822}          & {\textbf{0.898}} \\
{ML ROS 10\%}   & {0.579}          & {\textbf{0.697}} & {\textbf{0.959}} & {\textbf{0.828}} & {0.968}          & {0.759}          & {\textbf{0.898}} \\
\rowcolor[HTML]{EFEFEF} 
{ML ROS 20\%}   & {0.594}          & {0.676}          & {0.956}          & {0.816}          & {0.965}          & {0.751}          & {0.890}          \\
{ML ROS 30\%}   & {0.611}          & {0.675}          & {0.956}          & {0.816}          & {0.960}          & {0.754}          & {0.888}          \\
\rowcolor[HTML]{EFEFEF} 
{ML ROS 40\%}   & {\textbf{0.620}} & {0.694}          & {0.952}          & {0.823}          & {0.972}          & {\textbf{0.830}} & {0.897}          \\ \bottomrule

\end{tabular}
\caption{Comparison of results obtained by different resampling percentage of the ML ROS algorithm.}
\label{tab:ml_ros_comparison}
\end{table}

\begin{table}[h!]
\footnotesize
  \centering
  \setlength\tabcolsep{3pt}
\begin{tabular}{@{}lccccccc@{}}
\toprule
\multicolumn{1}{c} {Algorithm} & 
\begin{tabular}{c}ML \\ F1 \end{tabular} & 
\begin{tabular}{c}ML \\ mAP \end{tabular} & 
\begin{tabular}{c}ML \\ AUC \end{tabular} & \begin{tabular}{c}\textbf{ML} \\ \textbf{Score} \end{tabular} & \begin{tabular}{c}Bin \\ AUC \end{tabular} & 
\begin{tabular}{c}Bin \\ F1 \end{tabular}  & \begin{tabular}{c}\textbf{Model} \\ \textbf{Score} \end{tabular} \\ \midrule
\rowcolor[HTML]{EFEFEF} 
{No Resampling} & {0.595}                                           & {0.689}                                            & {0.957}                                            & {0.823}                                                       & {0.973}                                             & {\textbf{0.822}}                                   & {0.898}                                                          \\
\rowcolor[HTML]{FFFFFF} 
{LP ROS 10\%}   & {0.585}                                           & {\textbf{0.693}}                                   & {\textbf{0.962}}                                   & {\textbf{0.827}}                                              & {0.971}                                             & {0.778}                                            & {\textbf{0.899}}                                                 \\
\rowcolor[HTML]{EFEFEF} 
{LP ROS 20\%}   & {0.609}                                           & {0.684}                                            & {0.955}                                            & {0.820}                                                       & {\textbf{0.976}}                                    & {0.811}                                            & {0.898}                                                          \\
\rowcolor[HTML]{FFFFFF} 
{LP ROS 30\%}   & {\textbf{0.622}}                                  & {0.681}                                            & {0.949}                                            & {0.815}                                                       & {0.971}                                             & {0.800}                                            & {0.893}                                                          \\
\rowcolor[HTML]{EFEFEF} 
{LP ROS 40\%}   & {0.601}                                           & {0.660}                                            & {0.954}                                            & {0.807}                                                       & {0.967}                                             & {0.781}                                            & {0.887}                                                          \\ \bottomrule
\end{tabular}
\caption{Comparison of results obtained by different resampling percentage of the LP ROS algorithm.}
\label{tab:lp_ros_comparison}
\end{table}

The outcome of this investigation was that increasing the percentage of oversampling did not yield better results than using the 10\% resampling ratio baseline. Thus, using the LP ROS algorithm with a resampling ratio of 10\% resulted the best resampling strategy to reduce the effect of class imbalance in the model performance. Following experimentation employs the LP ROS resampling technique with a 10\% resampling ratio.

In the second experiment, a variety of popular loss functions designed for imbalanced datasets were tested. For this experiment, the mentioned model configuration integrating the best-performing resampling strategy was used, only changing the loss function to one of weighted binary cross-entropy (WBCE), binary cross-entropy (BCE), focal loss (FocalLoss), polynomial loss (PolyLoss), and asymmetric loss (ASL). Table \ref{tab:loss_comparison} shows the results obtained using different loss functions.



\begin{table}[h!]
\footnotesize
  \centering
  \setlength\tabcolsep{3pt}
\begin{tabular}{@{}rccccccc@{}}
\toprule
\multicolumn{1}{c} {Loss} & 
\begin{tabular}{c}ML \\ F1 \end{tabular} & 
\begin{tabular}{c}ML \\ mAP \end{tabular} & 
\begin{tabular}{c}ML \\ AUC \end{tabular} & \begin{tabular}{c}\textbf{ML} \\ \textbf{Score} \end{tabular} & \begin{tabular}{c}Bin \\ AUC \end{tabular} & 
\begin{tabular}{c}Bin \\ F1 \end{tabular}  & \begin{tabular}{c}\textbf{Model} \\ \textbf{Score} \end{tabular} \\ \midrule
\rowcolor[HTML]{EFEFEF} {WBCE}        & {0.562}          & {0.658}          & {0.951}          & {0.804}          & {0.947}          & {0.674}          & {0.876}          \\
{FocalLoss}   & {0.604}          & {0.680}          & {0.956}          & {0.818}          & {0.969}          & {0.819}          & {0.893}          \\
\rowcolor[HTML]{EFEFEF} {ASL}         & {\textbf{0.616}} & {0.688} & {0.958}          & {0.823}          & {0.969}          & {0.816}          & {0.896}          \\
{BCE}         & {0.595}          & {\textbf{0.689}} & {0.957}          & {0.823}          & {\textbf{0.973}} & {0.822}          & {\textbf{0.898}} \\
\rowcolor[HTML]{EFEFEF} {PolyLoss} & {0.599}          & {0.688} & {\textbf{0.960}} & {\textbf{0.824}} & {0.972} & {\textbf{0.832}} & {\textbf{0.898}} \\
\bottomrule
\end{tabular}
\caption{Comparison of results obtained using different loss functions.}
\label{tab:loss_comparison}
\end{table}

From the results obtained, it was concluded that the two best-performing loss functions were the conventional BCE and the PolyLoss. Both reached similar performance, but it was decided to continue further experiments using the PolyLoss function since it achieved better results when considering the rest of the calculated metrics.

Thus, it was concluded that the best strategy to alleviate the effect of class imbalance in the model's performance is to use the LP ROS resampling algorithm with a 10\% resampling rate altogether with the Polynomial Loss. Following experimentation makes use of this configuration.

As a final remark, from the results obtained, it was observed that most of the available techniques for alleviating the effect of the class imbalance present in multi-label datasets either did not help or marginally improved the results, which indicates that more methods and new ideas are in need in this field.

\subsubsection{Hyperparameter Optimization}
The final set of configuration-optimizing experiments focused on finding both the best-performing image size and batch size by following an incremental approach. To find the best-performing image size, different dimensions for the input images were selected and observed whether there was any difference in performance. Table \ref{tab:img_size_comparison} shows the results of this experiment.

\begin{table}[h!]
\footnotesize
  \centering
  \setlength\tabcolsep{3pt}
\begin{tabular}{@{}cccccccc@{}}
\toprule
\multicolumn{1}{c} {Image Size} & 
\begin{tabular}{c}ML \\ F1 \end{tabular} & 
\begin{tabular}{c}ML \\ mAP \end{tabular} & 
\begin{tabular}{c}ML \\ AUC \end{tabular} & \begin{tabular}{c}\textbf{ML} \\ \textbf{Score} \end{tabular} & \begin{tabular}{c}Bin \\ AUC \end{tabular} & 
\begin{tabular}{c}Bin \\ F1 \end{tabular}  & \begin{tabular}{c}\textbf{Model} \\ \textbf{Score} \end{tabular} \\ \midrule
\rowcolor[HTML]{EFEFEF} 
{384x384}                       & {0.585}          & {\textbf{0.693}} & {\textbf{0.962}} & {\textbf{0.827}} & {\textbf{0.971}} & {0.778}          & {\textbf{0.899}} \\
{448x448}                       & {0.579}          & {0.677}          & {0.953}          & {0.815}          & {0.970}          & {0.814}          & {0.892}          \\
\rowcolor[HTML]{EFEFEF} 
{512x512}                       & {0.580}          & {0.669}          & {0.952}          & {0.811}          & {0.964}          & {0.771}          & {0.887}          \\
{560x640}                       & {0.570}          & {0.683}          & {0.955}          & {0.819}          & {0.966}          & {0.834}          & {0.892}          \\
\rowcolor[HTML]{EFEFEF} 
{600x600}                       & {\textbf{0.619}} & {0.686}          & {0.959}          & {0.823}          & {0.960}          & {0.790}          & {0.891}          \\
{700x700}                       & {0.575}          & {0.674}          & {0.958}          & {0.816}          & {0.966}          & {\textbf{0.840}} & {0.891}          \\ \bottomrule
\end{tabular}
\caption{Comparison of results obtained by using different image sizes.}
\label{tab:img_size_comparison}
\end{table}

From the results obtained by using different image sizes, it was noticed that increasing the size of the images did not yield better results than the standard $384\times384$ size. A hypothesis for this result might be the high variability of resolution present in the dataset since each of the combined datasets uses a different image resolution which makes difficult to find a suitable size that does not increase or decrease too much the resolution of the images. The following experiment makes use of a $384\times384$ image size.

Next, to find the optimal batch size, an incremental approach was followed, starting with the base size of 16 and increasing it until reaching 64. Table \ref{tab:batch_size} shows the results of using different batch sizes.

\begin{table}[h!]
\footnotesize
  \centering
  \setlength\tabcolsep{3pt}
\begin{tabular}{@{}cccccccc@{}}
\toprule
{Batch Size} & {\begin{tabular}[c]{@{}c@{}}ML\\ F1\end{tabular}} & {\begin{tabular}[c]{@{}c@{}}ML\\ mAP\end{tabular}} & {\begin{tabular}[c]{@{}c@{}}ML\\ AUC\end{tabular}} & {\textbf{\begin{tabular}[c]{@{}c@{}}ML\\ Score\end{tabular}}} & {\begin{tabular}[c]{@{}c@{}}Bin\\ AUC\end{tabular}} & {\begin{tabular}[c]{@{}c@{}}Bin\\ F1\end{tabular}} & {\textbf{\begin{tabular}[c]{@{}c@{}}Model\\ Score\end{tabular}}} \\ \midrule
\rowcolor[HTML]{EFEFEF} 
{16}         & {0.585}                                           & {\textbf{0.693}}                                   & {\textbf{0.962}}                                   & {\textbf{0.827}}                                              & {0.971}                                             & {0.778}                                            & {0.899}                                                          \\
{32}         & {0.573}                                           & {0.685}                                            & {\textbf{0.962}}                                   & {0.824}                                                       & {\textbf{0.976}}                                    & {\textbf{0.824}}                                   & {\textbf{0.900}}                                                 \\
\rowcolor[HTML]{EFEFEF} 
{64}         & {\textbf{0.604}}                                  & {0.688}                                            & {0.959}                                            & {0.823}                                                       & {0.968}                                             & {0.813}                                            & {0.896}                                                          \\ \bottomrule
\end{tabular}
\caption{Comparison of results by using different batch sizes.}
\label{tab:batch_size}
\end{table}

From the comparison among different batch sizes, it can be noticed that increasing the size from 16 to 32 can achieve a slight gain in performance.

Following the aforementioned experiments, the optimal system configuration that maximizes the performance of the C-tran model on the MuReD dataset was determined. In summary, the C-Tran model with the Adam optimizer, a LR of $10^{-5}$, the Polynomial loss function, the DenseNet161 as feature extractor, the LP ROS algorithm with a 10\% resampling ratio, and an input image size of ($384 \times 384 \times 3$) is the best-performing configuration overall. Next, this configuration will be compared with other approaches for the problem of fundus multi-label classification task.

\subsection{Comparison with alternative approaches}
To compare the performance of the selected approach, previous works were identified, that tackled multi-label classification. Reproducibility was a significant factor since these approaches needed to be tested on the proposed MuReD dataset to have a fair comparison.
There were two main challenges when performing the comparison with state-of-the-art methods:

Most of the available research focused on multi-label classification using the ODIR dataset. However, this dataset uses patient-level diagnostics, i.e., images from both eyes are used to produce the final classification, and thus some of the published works developed specialized architectures to exploit this characteristic \cite{9098525, 9098340}, which in turn, makes them unsuitable for our dataset. Instead, the focus was placed on exploring the performance of techniques that were more flexible and could be used for classification using a single image, i.e., \cite{odir_vgg16, odir_effnetb3}. 
These techniques employ an ensemble of CNN models, i.e., using VGG16 and EfficientNetB3. Both approaches used the pre-trained weights from ImageNet and fine-tuned them on the ODIR dataset. Only \cite{odir_effnetb3} proposed the WBCE loss function to deal with the class imbalance and a pre-processing step using the CLAHE transformation \cite{clahe}.

Other works were also investigated, however, some of them were trained using large private datasets \cite{nature39}, causing their results to be difficult to reproduce on smaller datasets, or their method was difficult to replicate because of lack of code availability \cite{Cheng2021, 9349192}.

We also decided to compare with the winner of the RIADD challenge \cite{RIADD}, i.e., the competition where the RFMiD dataset was first introduced. The goal was to predict multiple diseases present in a fundus image. The winner of the competition proposed the use of an ensemble of EfficientNetB5 and B6 with different image sizes and a set of augmentations to increase the variability of the dataset.

To reproduce the selected approaches for evaluation on the MuReD dataset, each step and architecture design was followed as accurately as possible to the description presented by the authors. Indeed, there were instances, where some hyperparameters were not sufficiently detailed in the published articles. In those cases, trial and error experimentation was performed to determine the appropriate hyperparameter values to maximize the performance of the approach. Table \ref{tab:approach_comparison} shows the results of this comparison.


\begin{table}[h!]
\footnotesize
  \centering
  \setlength\tabcolsep{3pt}
\begin{tabular}{@{}lccccccc@{}}
\toprule
\multicolumn{1}{c} {Model} & 
\begin{tabular}{c}ML \\ F1 \end{tabular} & 
\begin{tabular}{c}ML \\ mAP \end{tabular} & 
\begin{tabular}{c}ML \\ AUC \end{tabular} & \begin{tabular}{c}\textbf{ML} \\ \textbf{Score} \end{tabular} & \begin{tabular}{c}Bin \\ AUC \end{tabular} & 
\begin{tabular}{c}Bin \\ F1 \end{tabular}  & \begin{tabular}{c}\textbf{Model} \\ \textbf{Score} \end{tabular} \\ \midrule
\rowcolor[HTML]{EFEFEF}Gour et al. \cite{odir_vgg16}                       & 0.010          & 0.262          & 0.825          & 0.544          & 0.872          & 0.507          & 0.708          \\
Wang et al. \cite{odir_effnetb3}                       & 0.315          & 0.379          & 0.845          & 0.612          & 0.897          & 0.669          & 0.754          \\
\rowcolor[HTML]{EFEFEF}RIADD $1^{st}$ \cite{hanson0910}                    & 0.208          & 0.380          & 0.881          & 0.630          & 0.893          & 0.637          & 0.762          \\
{Proposed model} & {\textbf{0.573}} & {\textbf{0.685}} & {\textbf{0.962}} & {\textbf{0.824}} & {\textbf{0.976}} & {\textbf{0.824}} & {\textbf{0.900}} \\ \bottomrule
\end{tabular}
\caption{Comparison of different proposed approaches for multi-label classification.}
\label{tab:approach_comparison}
\end{table}

From the results, it was observed that the C-Tran approach outperforms previous approaches, based on CNN architectures, by a considerable margin, demonstrating the superiority of the transformer-based method in the MuReD dataset.

To provide a better understanding of the performance of the C-Tran model, different performance metrics were calculated for each of the label classes. Table \ref{tab:class_scores} shows the obtained scores per class.

\begin{table}[h!]
\footnotesize
  \centering
  \setlength\tabcolsep{5pt}
\begin{tabular}{@{}ccccc@{}}
\toprule
Class  & Precision & Recall & F1    & AUC   \\ \midrule
\rowcolor[HTML]{EFEFEF} 
{DR}     & {0.859} & {0.859} & {0.859} & {0.962} \\
{NORMAL} & {0.865} & {0.786} & {0.824} & {0.976} \\
\rowcolor[HTML]{EFEFEF} 
{MH}     & {0.875} & {0.618} & {0.724} & {0.962} \\
{ODC}    & {0.661} & {0.750} & {0.703} & {0.966} \\
\rowcolor[HTML]{EFEFEF} 
{TSLN}   & {0.800} & {0.774} & {0.787} & {0.989} \\
{ARMD}   & {0.800} & {0.500} & {0.615} & {0.965} \\
\rowcolor[HTML]{EFEFEF} 
{DN}     & {0.708} & {0.531} & {0.607} & {0.938} \\
{MYA}    & {0.810} & {0.944} & {0.872} & {0.997} \\
\rowcolor[HTML]{EFEFEF} 
{BRVO}   & {0.929} & {0.813} & {0.867} & {0.994} \\
{ODP}    & {0.000} & {0.000} & {0.000} & {0.870} \\
\rowcolor[HTML]{EFEFEF} 
{CRVO}   & {0.600} & {0.545} & {0.571} & {0.981} \\
{CNV}    & {0.889} & {0.667} & {0.762} & {0.992} \\
\rowcolor[HTML]{EFEFEF} 
{RS}     & {1.000} & {0.545} & {0.706} & {0.971} \\
{ODE}    & {0.833} & {0.909} & {0.870} & {0.999} \\
\rowcolor[HTML]{EFEFEF} 
{LS}     & {0.500} & {0.556} & {0.526} & {0.990} \\
{CSR}    & {0.444} & {0.571} & {0.500} & {0.981} \\
\rowcolor[HTML]{EFEFEF} 
{HTR}    & {0.000} & {0.000} & {0.000} & {0.911} \\
{ASR}    & {0.000} & {0.000} & {0.000} & {0.971} \\
\rowcolor[HTML]{EFEFEF} 
{CRS}    & {0.400} & {0.333} & {0.364} & {0.988} \\
{OTHER}  & {0.587} & {0.519} & {0.551} & {0.851} \\ \bottomrule
\end{tabular}
\caption{Proposed model performance per class.}
\label{tab:class_scores}
\end{table}

From the results per class table, it was observed that the model had a good performance overall, with most of the AUC scores per class being above 90\%. There were only two classes where the model achieved a lower AUC score, i.e., the "OTHER" class, which is difficult to predict correctly since it is an "umbrella" class for many diseases that are not included in the original label set, and the "ODP" class since this disease manifests itself with subtle color changes in the optic disc \cite{odp}, which can be hard to detect correctly. For the "HTR" and "ASR" classes, there is a good AUC score but no results in the F1 score since the model is classifying these classes with low confidence, which does not meet the defined threshold of $0.5$.

\subsection{Class Activation Maps}
In this section, a visual interpretation is provided of the parts of the retinal image, that the model is focusing on when making predictions. This is achieved using the Class Activation Maps (CAMs) technique proposed by Zhou et al. \cite{zhou2015learning}. CAMs are generated by a weighted sum of the extracted visual patterns by the feature extractor. The weights are defined by the class-wise weights of the fully connected layer. The result of this weighted sum is then upsampled to the original image size as well as passed through a softmax or sigmoid output function.

\begin{figure*}[h!]
\centering
\includegraphics[scale=0.45]{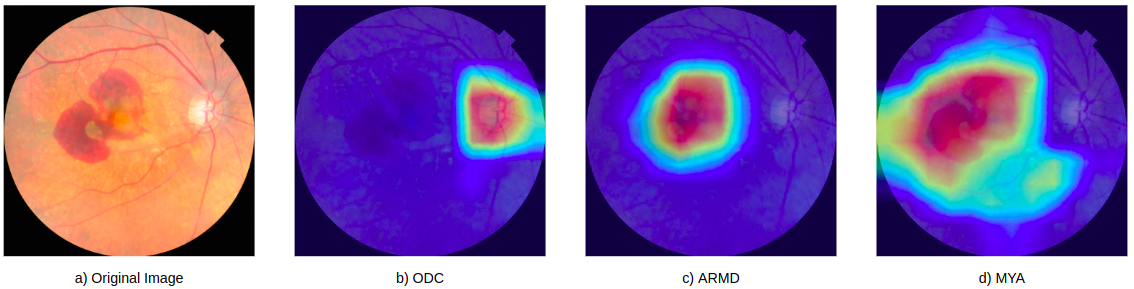}
\caption{a) The original image used for classification. b) Heat-map generated for Optic Disc Cupping, also known as Glaucoma. It is usually diagnosed by measuring the size of the cup in the optic nerve. The model gives a classification confidence of $99\%$ c) Heat-map for ARMD. The model focuses on the neovascular membrane present in the image, which is one of the main clinical characteristics, with a confidence of $98\%$ d) Heat-map for Myopia. The model focuses on the degenerative changes in the tissue surrounding the neovasculature, detecting it with a confidence of $100\%$}
\label{fig:cams}
\end{figure*}

CAMs were used as a visual explanation method to evaluate whether the model correctly learned the characteristics that distinguish various diseases. The Pytorch GradCam library \cite{jacobgilpytorchcam} implementation was used to generate heat maps of different predictions made by the C-Tran model. Figure \ref{fig:cams} shows a sample retinal image and the heat maps generated for each pathology present in the original image.

It is observed from the figure that the model is using completely different zones from the original image to classify each of the diseases, which supports the claim that it learned the different features associated with the occurrence of each disease.

\section{Conclusions}
\label{sec:conclusion}

In this work, a new dataset for the multi-label classification of retinal diseases on fundus images was created using three publicly available datasets and performing cleaning steps using an automatic metric for quality assessment and a minimum number of samples per class. The final version of this dataset contains 2208 samples for 20 different classes.

A novel pipeline for the multi-label classification of retinal diseases was proposed using for the first time a transformer-based model, performing a set of experiments to ensure the optimality of the configuration used, and comparing our approach with state-of-the-art techniques on the same task. The proposed method achieved superior results than previously proposed techniques.

In terms of future research, we will focus on finding more effective ways to deal with the class imbalance problem present in the proposed dataset, so as to improve model performance \cite{longtailed}. Another line of research is to bring the benefits obtained from the C-Tran architecture to different multi-label problems, within the medical imaging field.

Seeing the excellent results obtained in the multi-label fundus disease classification by transformer-based architectures, we envisage that this work motivates more research into integrating such architectures to more tasks in the medical field, not only in classification but a wide variety of tasks where more powerful models are needed.\label{sec:conc}

%
%
%

\ifCLASSOPTIONcaptionsoff
  \newpage
\fi



\bibliographystyle{unsrt}
\bibliography{mybibfile}
%



%

\vspace{-1cm}
\begin{IEEEbiography}[{\includegraphics[width=1in,height=1.25in,clip,keepaspectratio]{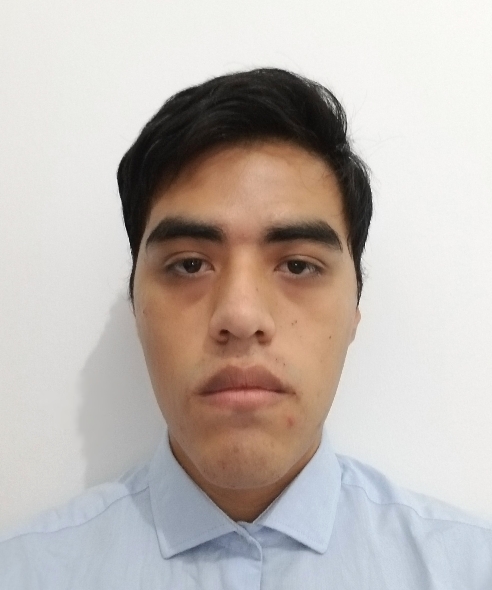}}]{Manuel A. Rodr\'{i}guez}
is a Computer Science M.Sc Student from the Department of Electrical Engineering and Computer Science at Khalifa University of Science and Technology. He received his Bachelor's degree in Artificial Intelligence from Universidad Panamericana, Aguascalientes, Mexico in 2020. His research interests include deep learning and computer vision. 
\end{IEEEbiography}
\vspace{-1cm}
\begin{IEEEbiography}[{\includegraphics[width=1in,height=1.25in,clip,keepaspectratio]{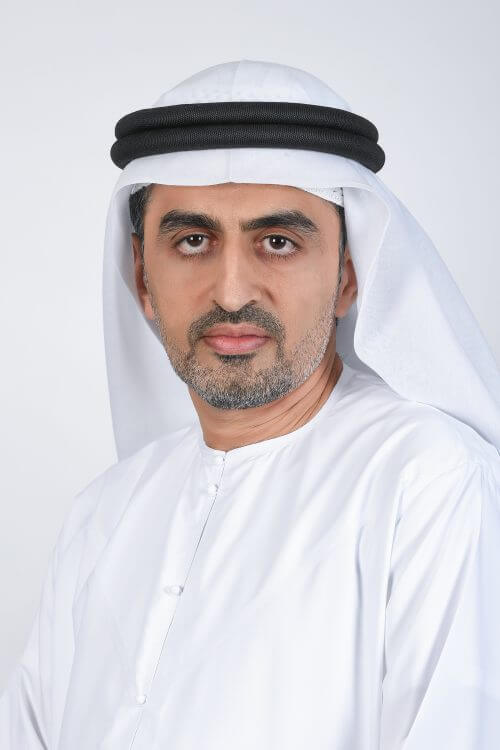}}]{Hasan AlMarzouqi}
is an Assistant Professor in the Department of Electrical Engineering and Computer Science at Khalifa University of Science and Technology. He received his Bachelor’s degree (with honors) and his M.Sc. degree, both in Electrical and Computer Engineering from Vanderbilt University, Nashville, Tennessee, in 2004 and 2006, respectively. He received his Ph.D. degree in Electrical and Computer Engineering from the Georgia Institute of Technology in 2014. Dr. Al-Marzouqi is a Senior Member of IEEE and a member of the IEEE Signal Processing Society. His current research interests include deep learning, artificial intelligence, digital rock physics, and bioinformatics.
\end{IEEEbiography}
\vspace{-1cm}
\begin{IEEEbiography}[{\includegraphics[width=1in,height=1.25in,clip,keepaspectratio]{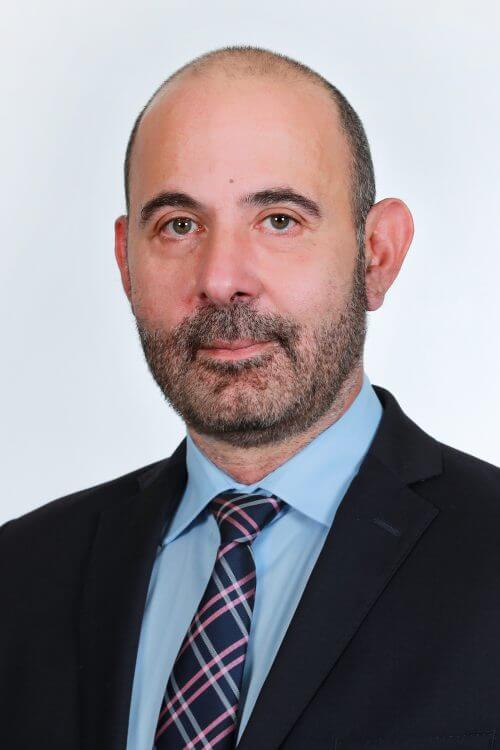}}]{Panos Liatsis}
is a Professor in the Department of Electrical Engineering and Computer Science at Khalifa University of Science and Technology. He received the Diploma in Electrical Engineering from the University of Thrace, Greece, and the Ph.D. in Electrical Engineering and Electronics from the University of Manchester, UK. He commenced his academic career at the University of Manchester, before joining City, University of London, UK, where he was a Professor and Head of the Electrical and Electronic Engineering Department. His research interests are image processing, computer vision, pattern recognition, and machine learning. 
\end{IEEEbiography}







\end{document}